\documentclass[a4paper,conference]{IEEEtran}
\usepackage{graphicx}
\usepackage{cite}

\ifCLASSINFOpdf
\else
\fi
\usepackage{float}
\usepackage{amsmath,bm}
\usepackage{graphicx}
\usepackage{adjustbox}
\begin{document}
%
\title{PaDiM: a Patch Distribution Modeling Framework for Anomaly Detection and Localization}

 \author{\IEEEauthorblockN{Thomas Defard, Aleksandr Setkov, Angelique Loesch, Romaric Audigier}
\IEEEauthorblockA{\textit{Université Paris-Saclay, CEA, List}, F-91120, Palaiseau, France\\
 thomas.defard@imt-atlantique.net, \{aleksandr.setkov, angelique.loesch, romaric.audigier\}@cea.fr\\}\\
}

\maketitle

\begin{abstract}
We present a new framework for Patch  Distribution Modeling, PaDiM, to concurrently detect and localize anomalies in images in a one-class learning setting. PaDiM makes use of a pretrained convolutional neural network (CNN) for patch embedding, and of multivariate Gaussian distributions to get a probabilistic representation of the normal class. It also exploits correlations between the different semantic levels of CNN to better localize anomalies. PaDiM outperforms current state-of-the-art approaches for both  anomaly detection and localization on the MVTec AD and STC datasets.
To match real-world visual industrial inspection, we extend the evaluation protocol to assess performance of anomaly localization algorithms on non-aligned dataset. 
The state-of-the-art performance and low complexity of PaDiM make it a good candidate for many industrial applications.


\end{abstract}


%
\IEEEpeerreviewmaketitle

\section{Introduction}

\begin{figure}[!t]

\includegraphics{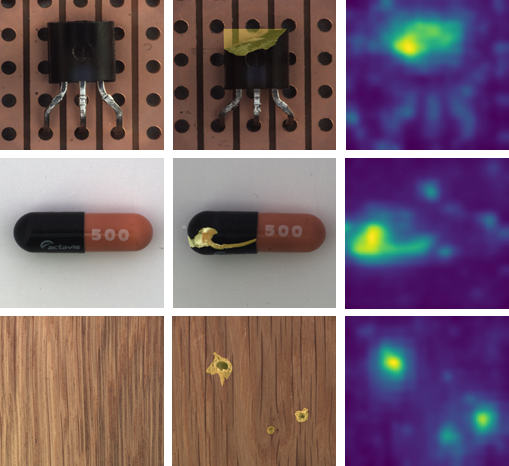}
\caption{Image samples from the MVTec AD \cite{bergmann2019mvtec}. \textit{Left column}: normal images of Transistor, Capsule and Wood classes. \textit{Middle column}: images of the same classes with the ground truth anomalies highlighted in yellow. \textit{Right column}: anomaly heatmaps obtained by our PaDiM model. Yellow areas correspond to the detected anomalies, whereas the blue areas indicate the normality zones. Best viewed in color.}



\label{fig:segim}
\end{figure}

Humans are able to detect heterogeneous or unexpected patterns in a set of homogeneous natural images. This task is known as anomaly or novelty detection and has a large number of applications, among which visual industrial inspections. However, anomalies are very rare events on manufacturing lines and cumbersome to detect manually. Therefore, anomaly detection automation would enable a constant quality control by avoiding reduced attention span and facilitating human operator work.
In this paper, we focus on anomaly detection and, in particular, on anomaly localization, mainly in an industrial inspection context. In computer vision, anomaly detection consists in giving an anomaly score to images. Anomaly localization is a more complex task which assigns each pixel, or each patch of pixels, an anomaly score to output an anomaly map. Thus, anomaly localization yields more precise and interpretable results. Examples of anomaly maps produced by our method to localize anomalies in images from the  MVTec Anomaly Detection (MVTec AD) dataset \cite{bergmann2019mvtec} are displayed in Figure \ref{fig:segim}.

Anomaly detection is a binary classification between the normal and the anomalous classes. However, it is not possible to train a model with full supervision for this task because we frequently lack anomalous examples, and, what is more, anomalies can have unexpected patterns. Hence, anomaly detection models are often estimated in a one-class learning setting,  \textit{i.e.}, when the training dataset contains only images from the normal class and anomalous examples are not available during the training. At test time, examples that differ from the normal training dataset are classified as anomalous. 


Recently, several methods have been proposed to combine anomaly localization and detection tasks in a one-class learning setting \cite{bergmann2019uninformed, venkataramanan2019attention, yi2020patch, cohen2020subimage}. However, either they require deep neural network training \cite{Bergman2020Classification-Based, venkataramanan2019attention} which might be cumbersome, or they use a K-nearest-neighbor (K-NN)  algorithm  \cite{knn} on the entire training dataset at test time \cite{yi2020patch, cohen2020subimage}. The linear complexity of the KNN algorithm increases the time and space complexity as the size of the training dataset grows. These two scalability issues may hinder the deployment of anomaly localization algorithms in industrial context.

To mitigate the aforementioned issues, we propose a new anomaly detection and localization approach, named
PaDiM for Patch Distribution Modeling. It makes use of a pretrained convolutional neural network (CNN) for embedding extraction and has the two following properties:  

 \begin{figure*}[!ht]

\caption{For each image patch corresponding to position $(i,j)$ in the largest CNN feature map, PaDiM learns the Gaussian parameters  $(\mu_{ij}, \Sigma_{ij})$ from the set of N training embedding vectors $X_{ij}=\{x_{ij}^k,k \in  [\![1,N  ]\!] \}$, computed from N different training images and three different pretrained CNN layers.}


\includegraphics[ width = 1\linewidth, height = 5.5cm] {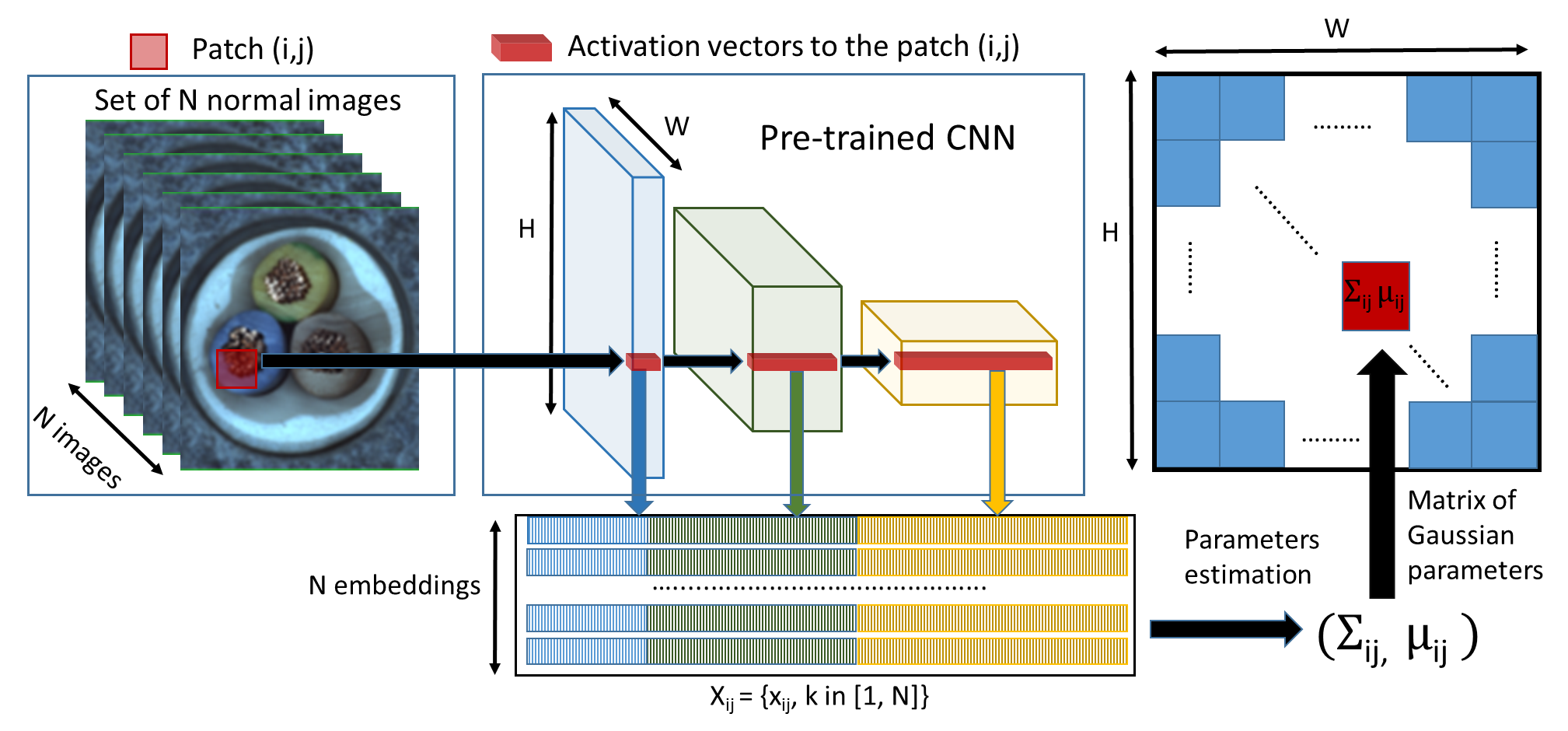}

\label{fig:overview-tr}
\end{figure*}
\begin{itemize}

  \item Each patch position is described by a multivariate Gaussian distribution;
  \item PaDiM takes into account the correlations between different semantic levels of a pretrained CNN.

\end{itemize}

With this new and efficient approach, PaDiM outperforms the existing state-of-the-art methods for anomaly localization and detection on the MVTec AD \cite{bergmann2019mvtec} and the ShanghaiTech Campus (STC) \cite{liu2018ano_pred} datasets. Besides, at test time, it has a low time and space complexity, independent of the dataset training size which is an asset for industrial applications. We also extend the evaluation protocol to assess model performance in more realistic conditions, \textit{i.e.}, on a non-aligned dataset.

\section{Related work}
Anomaly detection and localization methods can be categorized as either reconstruction-based or embedding similarity-based methods.

\textbf{Reconstruction-based methods} are widely-used for anomaly detection and localization. Neural network architectures like autoencoders (AE) \cite{Bergmann_2019, bergmann2019mvtec, gong2019memorizing, huang2019attribute}, variational autoencoders (VAE) \cite{kingma2013autoencoding, 8852144_vae, liu2019visually,venkataramanan2019attention} or generative adversarial networks (GAN) \cite{sabokrou2018adversarially, pidhorskyi2018generative, akcay2018ganomaly} are trained to reconstruct normal training images only. Therefore, anomalous images can be spotted as they are not well reconstructed. At the image level, the simplest approach is to take the reconstructed error as an anomaly score \cite{gong2019memorizing} but additional information from the latent space \cite{pidhorskyi2018generative, abati2018latent}, intermediate activations \cite{Kim2020RaPP} or a discriminator \cite{akay2019skipganomaly, akcay2018ganomaly} can help to better recognize anomalous images. Yet to localize anomalies, reconstruction-based methods can take the pixel-wise reconstruction error as the anomaly score \cite{bergmann2019mvtec} or the structural similarity \cite{Bergmann_2019}. Alternatively, the anomaly map can be a visual attention map generated from the latent space \cite{venkataramanan2019attention, liu2019visually}. Although reconstruction-based methods are very intuitive and interpretable, their performance is limited by the fact that AE can sometimes yield good reconstruction results for anomalous images too \cite{DBLP08550}.

\textbf{Embedding similarity-based methods} use deep neural networks to extract meaningful vectors describing an entire image for anomaly detection \cite{pmlr-v80-ruff18a, rippel2020modeling, bergman2020deep, Bergman2020Classification-Based} or an image patch for anomaly localization \cite{bergmann2019uninformed, cohen2020subimage, yi2020patch, Napoletano}. Still, embedding similarity-based methods that only perform anomaly detection give promising results but often lack interpretability  as it is not possible to know which part of an anomalous images is responsible for a high anomaly score. The anomaly score is in this case the distance between embedding vectors of a test image and reference vectors representing normality from the training dataset. The normal reference can be the center of a n-sphere containing embeddings from normal images \cite{pmlr-v80-ruff18a, yi2020patch}, parameters of Gaussian distributions \cite{rippel2020modeling, lee2018simple} or the entire set of normal embedding vectors \cite{bergman2020deep, cohen2020subimage}. The last option is used by SPADE \cite{cohen2020subimage} which has the best reported results for anomaly localization. However, it runs a K-NN algorithm on a set of normal embedding vectors at test time, so the inference complexity scales linearly to the dataset training size. This may hinder industrial deployment of the method. 

Our method, PaDiM, generates patch embeddings for anomaly localization, similar to the aforementioned approaches. However, the normal class in PaDiM is described through a set of Gaussian distributions that also model correlations between semantic levels of the used pretrained CNN model. Inspired by \cite{cohen2020subimage,rippel2020modeling}, we choose as pretrained networks a ResNet \cite{he2015deep}, a Wide-ResNet \cite{wdr50} or an EfficientNet \cite{efficientnet}. Thanks to this modelisation, PaDiM outperforms the current state-of-the-art methods. Moreover, its time complexity is low and independent of the training dataset size at the prediction stage.

\section{Patch Distribution Modeling}

\subsection{Embedding extraction}
\label{Embeddings extraction}

Pretrained CNNs are able to output relevant features for anomaly detection \cite{bergman2020deep}. Therefore, we choose to avoid ponderous neural network optimization by only using a pretrained CNN to generate patch embedding vectors.  The patch embedding process in PaDiM is similar to one from SPADE \cite{cohen2020subimage} and illustrated in Figure \ref{fig:overview-tr}. During the training phase, each patch of the normal images is associated to its spatially corresponding activation vectors in the pretrained CNN activation maps. Activation vectors from different layers are then concatenated to get embedding vectors carrying information from different semantic levels and resolutions, in order to encode fine-grained and global contexts.  As activation maps have a lower resolution than the input image, many pixels have the same embeddings and then form pixel patches with no overlap in the original image resolution. Hence, an input image can be divided in a grid of $(i,j) \in [1,W] \times[1,H]$ positions where  $W$x$H$ is the resolution of the largest activation map used to generate embeddings. Finally, each patch position  $(i,j)$ in this grid is associated to an embedding vector $x_{ij}$ computed as described above.

The generated patch embedding vectors may carry redundant information, therefore we experimentally  study the possibility  to reduce their size  (Section \ref{Ablative study}).  We noticed that randomly selecting few dimensions is more efficient that a classic Principal Component Analysis (PCA) algorithm \cite{doi:10.1080/14786440109462720}. This simple random dimensionality reduction significantly decreases the complexity of our model for both training and testing time while maintaining the state-of-the-art performance. Finally, patch embedding vectors from test images are used to output an anomaly map with the help of the learned parametric representation of the normal class described in the next subsection.
\subsection{Learning of the normality}
\label{subsection:my1}

To learn the normal image characteristics at position $(i,j)$, we first compute the set of patch embedding vectors at $(i,j)$,   $X_{ij}=\{x_{ij}^k,k \in  [\![1,N  ]\!] \}$ from the $N$ normal training images as shown on Figure \ref{fig:overview-tr}. To sum up the information carried by this set we make the assumption that $X_{ij}$ is generated by a multivariate Gaussian distribution $\mathcal{N}({\mu_{ij}},\boldsymbol{\Sigma_{ij}})$ where $\mu_{ij}$ is the sample mean of $X_{ij}$ and the sample covariance $\Sigma_{ij}$ is estimated as follows :
\begin{equation}
\Sigma_{ij} = {\frac{1}{N-1}}\sum_{k=1}^N (\mathbf{x_{ij}^k}-\mathbf{\mu_{ij}}) (\mathbf{x_{ij}^k}-\mathbf{\mu_{ij}})^\mathrm{T}  +\bm{\epsilon}I 
\label{eqn:sigma}
\end{equation}

where the regularisation term $\bm{\epsilon}I$ makes the sample covariance matrix $\Sigma_{ij}$ full rank and invertible. Finally, each possible patch position is associated with a multivariate Gaussian distribution as shown in Figure \ref{fig:overview-tr} by the matrix of Gaussian parameters.

Our patch embedding vectors carry information from different semantic levels. Hence, each estimated multivariate Gaussian distribution $\mathcal{N}({\mu_{ij}},\boldsymbol{\Sigma_{ij}})$ captures information from different levels too and $\boldsymbol{\Sigma_{ij}}$ contains the inter-level correlations. We experimentally show (Section \ref{Ablative study}) that modeling these relationships between the different semantic levels of the pretrained CNN helps to increase anomaly localization performance.

\subsection{Inference : computation of the anomaly map}
\label{inference}
Inspired by \cite{lee2018simple,rippel2020modeling}, we use the Mahalanobis distance \cite{mahala} $M(x_{ij})$ to give an anomaly score to the patch in position $(i,j)$ of a test image. $M(x_{ij})$ can be interpreted as   the distance between the test patch embedding $x_{ij}$ and learned distribution $\mathcal{N}({\mu_{ij}},\boldsymbol{\Sigma_{ij}})$, where $M(x_{ij})$ is computed as follows:

\begin{equation}
M(x_{ij}) = \sqrt{(x_{ij} - \mu_{ij})^T \Sigma_{ij}^{-1}(x_{ij} - \mu_{ij})}
\end{equation}
Hence, the matrix of Mahalanobis distances $M = (M(x_{ij}))_{1<i<W, 1<j<H}$ that forms an anomaly map can be computed. High scores in this map indicate the anomalous areas. The final anomaly score of the entire image is the maximum of anomaly map $M$. 
Finally, at test time, our method does not have the scalability issue of the K-NN based methods \cite{yi2020patch, cohen2020subimage, Napoletano, Bergman2020Classification-Based} as we do not have to compute and sort a large amount of distance values to get the anomaly score of a patch.
\newline

\section{Experiments}

\subsection{Datasets and  metrics}
\label{Dataset, metrics and experimental setup}

\textbf{Metrics}. To assess the localization performance we compute two threshold independent metrics. We use the Area Under the Receiver Operating Characteristic curve (AUROC) where the true positive rate is the percentage of pixels correctly classified as anomalous. Since the AUROC is biased in favor of large anomalies we also employ the per-region-overlap score (PRO-score) \cite{bergmann2019uninformed}. It consists in plotting, for each connected component, a curve of the mean values of the correctly classified pixel rates as a function of the false positive rate between 0 and 0.3. The PRO-score is the normalized integral of this curve. A high PRO-score means that both large and small anomalies are well-localized.

\textbf{Datasets}. We first evaluate our models on the MVTec AD \cite{bergmann2019mvtec} designed to test anomaly localization algorithms for industrial quality control and in a one-class learning setting. It contains 15 classes of approximately 240 images. The original image resolution is between 700x700 and 1024x1024. There are 10 object and 5 texture classes. Objects are always well-centered and aligned in the same way across the dataset as we can see in Figure \ref{fig:segim} for classes Transistor and Capsule. In addition to the original dataset, to assess performance of anomaly localization models in a more realistic context, we create a modified version of the MVTec AD, referred as  Rd-MVTec AD, where we apply random  rotation  (-10,  +10)  and  random  crop  (from  256x256 to 224x224) to both the train and test sets. This modified version of the MVTec AD may better describe  real  use  cases  of  anomaly  localization  for  quality control  where  objects  of  interest  are not always centered  and aligned in the image. 

For further evaluation, we also test PaDiM on the Shanghai Tech Campus (STC) Dataset \cite{liu2018ano_pred} that simulates video surveillance from a static camera. It contains 274 515 training and 42 883 testing frames divided in 13 scenes. The original image resolution is 856x480. The training  videos are composed of normal sequences and test videos have anomalies like the presence of vehicles in pedestrian areas or people fighting.

\subsection{Experimental setups}
\label{experimental setup}
We train PaDiM with different backbones, a ResNet18 (R18) \cite{he2015deep}, a Wide ResNet-50-2 (WR50) \cite{wdr50} and an EfficientNet-B5 \cite{efficientnet}, all pretrained on ImageNet \cite{imagenet_cvpr09}. Like in \cite{cohen2020subimage}, patch embedding vectors are extracted from the first three layers when the backbone is a ResNet, in order to combine information from different semantic levels, while keeping a high enough resolution for the localization task. Following this idea, we extract patch embedding vectors from layers 7 (level 2), 20 (level 4), and 26 (level 5), if an EfficientNet-B5 is used. We also apply a random dimensionality reduction (Rd) (see Sections \ref{Embeddings extraction} and \ref{Ablative study}). Our model names indicate the backbone and the dimensionality reduction method used, if any. For example, PaDiM-R18-Rd100 is a PaDiM model  with a ResNet18 backbone using 100 randomly selected dimensions for the patch embedding vectors. By default we use $\bm{\epsilon} = 0.01$ for  the $\bm{\epsilon}$ from Equation~\ref{eqn:sigma}.

We reproduce  the model SPADE \cite{cohen2020subimage} as described in the original publication with a Wide ResNet-50-2 (WR50) \cite{wdr50} as backbone.  For SPADE and PaDiM we apply the same prepocessing as in \cite{cohen2020subimage}. We resize the images from the MVTec AD to 256x256 and center crop them to 224x224. For the images from the STC we use a 256x256 resize only. We resize the images and the localization maps using bicubic interpolation and we use a Gaussian filter on the anomaly maps with parameter $\sigma = 4$ like in \cite{cohen2020subimage}.

We also implement our own VAE as a reconstruction-based baseline implemented with a ResNet18 as encoder and  a 8x8 convolutional latent variable. It is trained on each MVTec AD class with 10 000 images using the following data augmentations operations:  random rotation ($-2^{\circ}$, $+2^{\circ}$), 292x292 resize, random crop to 282x282, and finally center crop to 256x256. The training is performed during 100 epochs with the Adam optimizer \cite{kingma2013autoencoding} with an initial learning rate of $10^{-4}$ and a batch size of 32 images. The anomaly map for the localization corresponds to the pixel-wise L2 error for reconstruction.

\section{Results}
\subsection{Ablative studies}
First, we evaluate the impact of modeling correlations between semantic levels in PaDiM and explore the possibility to simplify our method through dimensionality reduction.
\label{Ablative study}

\textbf{Inter-layer correlation}. The combination of Gaussian modeling and the Mahalanobis distance has already been employed in previous works to detect adversarial attacks \cite{lee2018simple} and for anomaly detection \cite{rippel2020modeling} at the image level. However those methods do not model correlations between different CNN's semantic levels as we do in PaDiM. In Table \ref{semantic level} we show the anomaly localization performance on the MVTec AD of PaDiM with a ResNet18 backbone when using only one of the first three layers (Layer 1, Layer 2, or Layer 3) and  when summing the outputs of these 3 models to form an ensemble method that takes into account the first three layers but not the correlations between them (Layers 1+2+3). The last row of Table \ref{semantic level} (PaDiM-R18) is our proposed version of PaDiM where each patch location is described by one Gaussian distribution taking into account the first three ResNet18 layers and correlations between them.
It can be observed that using Layer 3 produces the best results in terms of AUROC among the three layers. It is due to the fact that Layer 3 carries higher semantic level information which helps to better describe normality. However, Layer 3 has a slightly worse PRO-score than Layer 2 that can be explained by the lower resolution of Layer 2 which affects the accuracy of anomaly localization. As we see in the two last rows of Table \ref{semantic level}, aggregating information from different layers can solve the trade-off issue between high semantic information and high resolution. Unlike model Layer 1+2+3 that simply sums the outputs, our model PaDiM-R18 takes into account correlations between semantic levels. As a result, it outperforms Layer 1+2+3 by 1.1p.p (percent point) for AUROC and 1.8p.p for PRO-score. It confirms the relevance of modeling correlation between semantic levels. 
\begin{table}[!t]
\caption{Study of the anomaly localization performance using different semantic-level CNN layers. Results are displayed as tuples (AUROC\%, PRO-score\%)  on the MVTec AD.}
\begin{center}

  \resizebox{0.5\textwidth}{!}{{}
\begin{tabular}{|p{1.9cm}|c|c|c|} 
\hline
 Layer used
 &all texture classes
 &all object classes
 & all classes \\ \hline
Layer 1&  (93.1, 87.1)& (95.6, 86.5)&  (94.8, 86.8) \\ \hline
Layer 2&  (95.0, 89.7)& (96.1, 87.9)& (95.7, 88.5) \\ \hline
Layer 3&  (94.8, 89.6)& (97.1, 87.7)& (95.7, 88.3) \\ \hline
Layer 1+2+3&  (95.4, 90.7)& (96.3, 88.1)& (96.0, 89.0) \\ \hline

PaDiM-R18&  (\textbf{96.3}, \textbf{92.3})& (\textbf{97.5}, \textbf{90.1})& (\textbf{97.1}, \textbf{90.8})\\ \hline
\end{tabular}}
\end{center}

\label{semantic level}
\end{table}
\begin{table}[h]
  \centering
\caption{Study of the anomaly localization performance with a dimensionality reduction from 448 to 100 and 200 using PCA or random feature selection (Rd). Results are displayed as tuples (AUROC\%, PRO-score\%)  on the MVTec AD.}
   \resizebox{0.5\textwidth}{!}{{}
\begin{tabular}{|c|c|c|c|} 
\hline
&all texture classes &all object  classes& all classes \\ \hline\hline
Rd 100&  (95.7, 91.3)& (97.2, 89.4)&  (96.7, 90.5) \\
PCA 100&  (93.7, 88.9)& (93.5, 84.1)& (93.5, 85.7)\\ \hline\hline
Rd 200&  (96.1, 92.0)& (97.5, 89.8)& (97.0, 90.5)\\ 
PCA 200&  (95.1, 91.8)& (96.0, 88.1)& (95.7, 89.3)\\ \hline\hline
all (448)
 &  (\textbf{96.3}, \textbf{92.3})
 & (\textbf{97.5}, \textbf{90.1})
 & (\textbf{97.1}, \textbf{90.8})\\ \hline
\end{tabular}
}

\label{reduction}
\end{table}
\begin{table*}[!ht]
\caption{Comparison of our PaDiM models with the state-of-the-art for the anomaly localization on the MVTec AD. Results are displayed as tuples (AUROC\%, PRO-score\%)}

   \resizebox{1\textwidth}{!}{

\begin{tabular}{|p{2.3cm}||p{1.3cm}|p{1.2cm}|p{1.4cm}||p{1.2cm}|p{1.3cm}|p{1.4cm}||p{1.4cm}|p{1.6cm}|}
\hline
\multicolumn{1}{|c|}{Type} &\multicolumn{3}{c|}{Reconstruction-based methods} & \multicolumn{3}{c|}{Embedding similarity based methods} & \multicolumn{2}{c|}{Our methods}\\
\hline

Model  & AE simm \cite{Bergmann_2019, bergmann2019mvtec, bergmann2019uninformed} &  AE L2 \cite{bergmann2019mvtec, bergmann2019uninformed}& VAE & Student \cite{bergmann2019uninformed}  & Patch SVDD \cite{yi2020patch}& SPADE \cite{cohen2020subimage} &  PaDiM-R18-Rd100 & PaDiM-WR50-Rd550 \\ \hline

Carpet & (87, 64.7)  &  (59, 45.6) &(59.7, 61.9)& (-, 69.5) & (92.6, -) & (97.5, 94.7) &(98.9, 96.0) & (\textbf{99.1}, \textbf{96.2}) \\ 
Grid & (94, 84.9) & (90, 58.2) &(61.2, 40.8)& (-, 81.9)  & (96.2, -) & (93.7, 86.7)  & (94.9, 90.9)  &(\textbf{97.3}, \textbf{94.6}) \\
Leather & (78, 56.1)  & (75, 81.9) &(67.1, 64.9)& (-, 81.9) &(97.4, -) & (97.6, 97.2)  & (99.1, 97.9) & (\textbf{99.2},  \textbf{97.8})\\ 
Tile & (59, 17.5)  & (51, 89.7) &(51.3, 24.2) & (-, 91.2) &(91.4, -) & (87.4, 75.9) & (91.2, 81.6) &(\textbf{94.1}, \textbf{86.0})  \\  
Wood & (73, 60.5) & (73, 72.7) & (66.6, 57.8) & (-, 72.5)  &(90.8, -) & (88.5, 87.4) & (93.6, 90.3)  & (\textbf{94.9}, \textbf{91.1}) \\ \hline
All texture classes & (78, 56.7) &  (70, 69.6)  &(61.2, 49.9)& (-, 79.4) &(93.7, -) & (92.9, 88.4) &(95.6, 91.3) & (\textbf{96.9}, \textbf{93.2}) \\ \hline 
Bottle & (93, 83.4)  & (86, 91.0)  &(83.1, 70.5)& (-, 91.8) &(98.1, -)&  (\textbf{98.4}, \textbf{95.5}) & (98.1, 93.9) & (98.3, 94.8) 
\\ Cable & (82, 47.8) & (86, 82.5) & (83.1, 77.9)&(-, 86.5) &(96.8, -) & (\textbf{97.2}, \textbf{90.9}) & (95.8, 86.2)& (96.7, 88.8)\\
Capsule  & (94, 86.0) & (88, 86.2) &(81.7, 77.9)& (-, 91.6) &(95.8, -) &  (\textbf{99.0} ,\textbf{93.7}) & (98.3, 91.9) & (98.5, 93.5)\\ 
Hazelnut & (97, 91.6) & (95, 91.7) &(87.7, 77.0)& (-, 93.7)& (97.5, -) &  (\textbf{99.1}, \textbf{95.4}) & (97.7, 91.4)& (98.2, 92.6)\\ 
Metal Nut& (89, 60.3) & (86, 83.0) &(78.7, 57.6)& (-, 89.5) &(98.0, -) & (\textbf{98.1}, \textbf{94.4})& (96.7, 81.9)& (97.2, 85.6)\\
Pill & (91, 83.0) & (85, 89.3) &(81.3, 79.3)& (-, 93.5) & (95.1, -) &   (\textbf{96.5}, \textbf{94.6}) & (94.7, 90.6) &  (95.7, 92.7)\\
Screw & (96, 88.7) & (96, 75.4) & (75.3, 66.4) &(-, 92.8) &(95.7, -)&   (\textbf{98.9}, \textbf{96.0})&(97.4, 91.3)& (98.5, 94.4)\\  
Toothbrush & (92, 78.4) & (93, 82.2) &(91.9, 85.4)& (-, 86.3) & (98.1, -)&  (97.9, \textbf{93.5})&(98.7, 92.3) &(\textbf{98.8}, 93.1)\\ 
Transistor & (90, 72.5) & (86, 72.8) &(75.4, 61.0)& (-, 70.1) & (97.0, -) &  (94.1, \textbf{87.4}) &(97.2, 80.2)  &(\textbf{97.5}, 84.5) \\  
Zipper & (88, 66.5) & (77, 83.9) &(71.6, 60.8)& (-, 93.3) & (95.1, -) &(96.5, 92.6) &(98.2, 94.7) &  (\textbf{98.5}, \textbf{95.9})  \\ \hline
All object classes & (91, 75.8) & (88, 83.8) &(81.0, 71.4) &(-, 88.9) & (96.7, -) & (97.6, \textbf{93.4}) &(97.3, 89.4)  & (\textbf{97.8}, 91.6) \\ \hline
All classes & (87, 69.4) & (82, 79.0) &(74.4, 64.2) & (-, 85.7) & (95.7, -) &(96.5, 91.7) &(96.7, 90.1) &(\textbf{97.5}, \textbf{92.1}) \\ \hline
\end{tabular}
}

\label{tab:tab_locmvtec}
\end{table*}

\textbf{Dimensionality reduction}. PaDiM-R18 estimates multivariate Gaussian distributions from sets of patch embeddings vectors of 448 dimensions each. Decreasing the embedding vector size would reduce the computational and memory complexity of our model. We study two different dimensionality reduction methods. The first one consists in applying a Principal Component Analysis (PCA) algorithm to reduce the vector size to 100 or 200 dimensions. The second method is a random feature selection where we randomly select features before the training. In this case, we train 10 different models and take the average scores.  Still the randomness does not change the results between different seeds as the standard error mean (SEM) for the average AUROC is always between $10^{-4}$ and $10^{-7}$.

 From Table \ref{reduction} we can notice that for the same number of dimensions, the random dimensionality reduction (Rd) outperforms the PCA on all the MVTec AD classes by at least 1.3p.p in the AUROC and 1.2p.p in the PRO-score. It can be explained by the fact that PCA selects the dimensions with the highest variance which may not be the ones that help to discriminate the normal class from the anomalous one \cite{rippel2020modeling}.
 It can also be noted from Table \ref{reduction} that randomly reducing the embedding vector size to only 100 dimensions has a very little impact on the anomaly localization performance. The results drop only by 0.4p.p in the AUROC and 0.3p.p in the PRO-score. This simple yet effective dimensionality reduction method significantly reduces PaDiM time and space complexity as it will be shown in Section \ref{PaDiM Scalability gain}.

\subsection{Comparison with the state-of-the-art}
\label{Comparison with the state-of-the-art}
\textbf{Localization}. In Table \ref{tab:tab_locmvtec}, we show the AUROC and the PRO-score results for anomaly localization on the MVTec AD. For a fair comparison, we used a Wide ResNet-50-2 (WR50) as this backbone is used in SPADE \cite{cohen2020subimage}. Since the other baselines have smaller backbones, we also try a ResNet18 (R18). We randomly reduce the embedding size to 550 and 100 for PaDiM with WR50 and R18 respectively.

We first notice that PaDiM-WR50-Rd550 outperforms all the other methods in both the PRO-score and the AUROC on average for all the classes. PaDiM-R18-Rd100 which is a very light model also outperforms all models in the average AUROC on the MVTec AD classes by at least 0.2p.p. When we further analyze the PaDiM performances, we see that the gap for the object classes is small as PaDiM-WR50-Rd550 is the best only in the AUROC (+0.2p.p) but SPADE \cite{cohen2020subimage} is the best in the PRO-score (+1.8p.p). However, our models are particularly accurate on texture classes. PaDiM-WR50-Rd550 outperforms the second best model SPADE \cite{cohen2020subimage} by 4.8p.p and 4.0p.p in the PRO-score and the AUROC respectively on average on texture classes. Indeed, PaDiM learns an explicit probabilistic model of the normal classes contrary to SPADE \cite{cohen2020subimage} or Patch-SVDD \cite{yi2020patch}. It is particularly efficient on texture images because even if  they are not aligned and centered like object images, PaDiM effectively captures their statistical similarity accross the normal train dataset.

Additionally, we evaluate our model on the STC dataset. We compare our method to the two best reported models performing anomaly localization without temporal information, CAVGA-RU \cite{venkataramanan2019attention} and SPADE \cite{cohen2020subimage}.
\begin{table}[b!]
\caption{Comparison of our PaDiM model with the state-of-the-art for the anomaly localization on the STC in the AUROC\%.}
\centering

\label{tab:stc}
\resizebox{0.5\textwidth}{!}{
\begin{tabular}{|c|c|c|c|}
\hline


Model &  CAVGA-RU \cite{venkataramanan2019attention} & SPADE \cite{cohen2020subimage}& PaDiM-R18-Rd100 \\ \hline

AUROC score\% & 85 & 89.9 & \textbf{91.2}  \\
 \hline
\end{tabular}
}

\label{tab:rocmstc}
\end{table}
As shown in Table \ref{tab:rocmstc}, the best result (AUROC) on the STC dataset is achieved with our simplest model PaDiM-R18-Rd100 by a 2.1p.p. margin. In fact, pedestrian positions in images are highly variable in this dataset and, as shown in Section \ref{PaDiM on non-aligned dataset}, our method performs well on non-aligned datasets.


\begin{table*}[t]
\caption{Anomaly detection results (at the image level) on the MVTec AD using AUROC\%.}
\centering
\resizebox{1\textwidth}{!}{
\begin{tabular}{|c|p{1.2cm}|p{0.9cm}|p{1.3cm}|p{1.3cm}|p{2.4cm}|p{2cm}|p{2.3cm}|}
\hline


Model  &GANomaly \cite{akay2019skipganomaly} & ITAE \cite{huang2019attribute} & Patch SVDD \cite{yi2020patch} & SPADE\cite{cohen2020subimage} (WR50) & MahalanobisAD\cite{rippel2020modeling} (EfficientNet-B4)  & PaDiM-WR50-Rd550& PaDiM EfficientNet-B5\\ \hline
all textures classes& - & - &  94.6 & - & 97.2 & 98.8 & \textbf{99.0}  \\

all objects classes& - & - &  90.9 & - & 94.8 & 93.6 &\textbf{97.2} \\
all classes &76.2 & 83.9&  92.1 & 85.5 & 95.8 & 95.3 & \textbf{97.9}  \\
 \hline
\end{tabular}
}

\label{tab:rocdet}
\end{table*}

\textbf{Detection}. By taking the maximum score of the anomaly maps issued by our models (see Section  \ref{inference}) we give anomaly scores to entire images to perform anomaly detection at the image level. We test PaDiM for anomaly detection with a Wide ResNet-50-2 (WR50) \cite{wdr50} used in SPADE and an EfficientNet-B5 \cite{efficientnet}. The Table \ref{tab:rocdet} shows that our model PaDiM-WR50-Rd550 outperforms every method except MahalanobisAD \cite{rippel2020modeling} with their best reported backbone, an EfficientNet-B4. Still our PaDiM-EfficientNet-B5 outperforms every model by at least 2.6p.p on average on all the classes in the AUROC. Besides, contrary to the second best method for anomaly detection, MahalanobisAD \cite{rippel2020modeling}, our model also performs anomaly segmentation which characterizes more precisely the anomalous areas in the images.

\label{Comparison with the state-of-the-art}

\subsection{Anomaly localization on a  non-aligned dataset}
\label{PaDiM on non-aligned dataset}
To estimate the robustness of anomaly localization methods, we train and evaluate the performance of PaDiM and several state-of-the-art methods (SPADE \cite{cohen2020subimage}, VAE) on a modified version of the MVTec AD, Rd-MVTec AD, described in Section \ref{Dataset, metrics and experimental setup}.
Results of this experiment are displayed in Table \ref{tab:modifiedmvtec}. 
For each test configuration we run 5 times data preprocessing on the MVTec AD with random seeds to obtain 5 different versions of the dataset, denoted as Rd-MVTec AD. Then, we average the obtained results and report them in Table \ref{tab:modifiedmvtec}.
According to the presented results, PaDiM-WR50-Rd550 outperforms the other models on both texture and object classes in the PRO-score and the AUROC. Besides, the SPADE \cite{cohen2020subimage} and VAE performances on the Rd-MVTec AD decrease more than the performance of PaDiM-WR50-Rd550 when comparing to the results obtained on the normal MVTec AD (refer to Table \ref{tab:tab_locmvtec}). The AUROC results decrease by 5.3p.p for PaDiM-WR50-Rd550 against 12.2p.p and 8.8p.p decline for VAE and SPADE respectively. Thus, we can conclude that our method seems to be more robust to non-aligned images than the other existing and tested works. 
\begin{table}[H]
\centering
\caption{Anomaly localization results on the non-aligned Rd-MVTec AD. Results are displayed as tuples (AUROC\%, PRO-score\%)}
\resizebox{0.5\textwidth}{!}{
\begin{tabular}{|c||p{1.5cm}||p{1.4cm}|p{1.8cm}||}
\hline
 Model & VAE (R18) &  SPADE (WR50) & PaDiM-WR50-Rd550    \\ \hline

all texture classes  &(54.7, 23.1) & (84.6, 75.6) &  (\textbf{92.4}, \textbf{77.9})\\ \hline
all object classes &(65.8, 30.2) & (88.2, 65.8)   & (\textbf{92.1}, \textbf{70.8})  \\ \hline
all classes & (62.1, 27.8) & (87.2, 69.0) &(\textbf{92.2}, \textbf{73.1}) \\\hline
\end{tabular}

}

\label{tab:modifiedmvtec}
\end{table}

\subsection{Scalability gain}
\label{PaDiM Scalability gain}
\begin{table}[b!]
\centering
\caption{Average inference time of anomaly localization in seconds on the MVTec AD with a CPU intel i7-4710HQ @ 2.50GHz.}
\resizebox{0.5\textwidth}{!}{
\begin{tabular}{|c|p{0.9cm}|p{0.8cm}|p{1.4cm}|p{1.7cm}|}
\hline
Model & SPADE (WR50) & VAE (R18) & PaDiM R18-Rd100 &PaDiM-WR50-Rd550  \\ \hline
Inference time (sec.)  & 7.10 & 0.21 & 0.23 & 0.95\\ \hline

\end{tabular}
}

\label{tab:inference_time}
\end{table}


\textbf{Time complexity}. In PaDiM, the training time complexity scales linearly with the dataset size because the Gaussian parameters are estimated using the entire training dataset. However, contrary to the methods that require to train deep neural networks, PaDiM uses a pretrained CNN, and, thus, no deep learning training is required which is often a complex procedure. Hence, it is very fast and easy to train it on small datasets like MVTec AD. For our most complex model PaDiM-WR50-Rd550, the training on a CPU (Intel CPU 6154 3GHz 72th) with a serial implementation takes on average 150 seconds on the MVTec AD classes and 1500 seconds on average on the STC video scenes. These training procedures could be further accelerated using GPU hardware for the forward pass and the covariance estimation. In contrast, training the VAE with 10 000 images per class on the MVTec AD following the procedure described in Section \ref{experimental setup} takes  2h40 per class using one GPU NVIDIA P5000. Conversely, SPADE \cite{cohen2020subimage} requires no training as there are no parameters to learn. Still, it computes and stores in the memory before testing all the embedding vectors of the normal training images. Those vectors are the inputs of a K-NN algorithm which makes SPADE's inference very slow as shown in Table \ref{tab:inference_time}.

In Table \ref{tab:inference_time}, we measure the model inference time using a mainstream CPU (Intel i7-4710HQ CPU @ 2.50GHz) with a serial implementation.
On the MVTec AD, the inference time of SPADE is around seven times slower than our PaDiM model with equivalent backbone because of the computationally expensive NN search. Our VAE implementation, which is similar to most reconstruction-based models, is the fastest model but our simple model PaDiM-R18-Rd100 has the same order of magnitude for the inference time. While having similar complexity, PaDiM largely outperfoms the VAE methods (see Section \ref{Comparison with the state-of-the-art}). 

\textbf{Memory complexity}. Unlike SPADE \cite{cohen2020subimage} and Patch SVDD \cite{yi2020patch}, the space complexity of our model is independent  of the dataset training size and depends only on the image resolution. PaDiM  keeps in the memory only the pretrained CNN and the Gaussian parameters associated with each patch. In Table \ref{tab:memory} we show the memory requirement of SPADE, our VAE implementation, and PaDiM, assuming that parameters are encoded in float32.
Using equivalent backbone, SPADE has a lower memory consumption than PaDiM on the MVTec AD. However, when using SPADE on a larger dataset like the STC, its memory consumption becomes intractable, whereas  PaDiM-WR50-Rd550 requires seven times less memory. The PaDiM space complexity increases from the MVTec AD to the STC only because the input image resolution is higher in the latter dataset as described in Section \ref{experimental setup}.
Finally, one of the advantages of our framework PaDiM is that the user can easily adapt the method by choosing the backbone and the embedding size to fit its inference time requirements, resource limits, or expected performance.

\begin{table}[!ht]
\centering
\caption{Memory requirement in Gb of the anomaly localization methods trained on the MVTec AD and the STC dataset. }
\resizebox{0.5\textwidth}{!}{
\begin{tabular}{|c|p{1.4cm}|p{1cm}|p{1.4cm}|p{1.6cm}|}
\hline
model & SPADE
(WR50)& VAE (R18) & PaDiM R18-Rd100 &PaDiM-WR50-Rd550  \\ \hline
MVTec AD & 1.4 & 0.09 &  0.17& 3.8  \\ \hline
STC &  37.0 & - & 0.21 & 5.2  \\ \hline
\end{tabular}
}

\label{tab:memory}
\end{table}

\section{Conclusion}We  have  presented  a framework called PaDiM for  anomaly detection and localization in one-class learning setting which is based on distribution modeling. It achieves state-of-the-art performance on MVTec AD and STC datasets.
Moreover, we extend the evaluation  protocol to non-aligned data and the first results show that PaDiM can be robust on these more realistic data.
PaDiM low memory and time consumption and its ease of use make it suitable for various applications, such as visual industrial control. 
\bibliographystyle{IEEEtran}

\bibliography{IEEEabrv, references}

\end{document}